\documentclass[runningheads]{llncs}
\usepackage[T1]{fontenc}

\usepackage{booktabs}
\usepackage{multirow}
\usepackage{adjustbox}
\usepackage{graphicx,verbatim}
\usepackage{graphicx}
\usepackage{caption} 

\usepackage{amsmath}
\usepackage{amssymb}
\usepackage{booktabs}
\usepackage{array}
\usepackage{makecell}
\usepackage{color}
\usepackage{multirow}
\usepackage{tabularx}
\usepackage{hyperref}
\usepackage{url}
\usepackage[table,xcdraw]{xcolor}
\usepackage{amsmath}
\usepackage{xcolor}
\usepackage{graphicx} 
\usepackage{booktabs}
\usepackage[misc]{ifsym}

\begin{document}
\titlerunning{Blind-sweep Ultrasound Fetal Birth Weight Estimation}
\title{Foundation Model-driven Key Anatomy Frame Selection for Blind-sweep Ultrasound Fetal Birth Weight Estimation}
%

\author{
Le Ou\inst{1,2}\thanks{Le Ou and Xiliang Zhu contributed equally.}\and 
Xiliang Zhu\inst{3\star} \and
Huanwen Liang\inst{1,2} \and
Wenxiong Pan\inst{3} \and \\
Yuhao Huang\inst{1,4} \and
Yuxiang Deng\inst{3} \and
Xuan Sheng\inst{3,5} \and
Hong Yin\inst{5} \and
Juhua Xiao\inst{6} \and \\
Xin Zhou\inst{6}\textsuperscript{(\Letter)} \and
Dong Ni\inst{1,3,7}\textsuperscript{(\Letter)}
}


\authorrunning{L. Ou et al.}

\institute{
Medical Ultrasound Image Computing (MUSIC) Lab, Shenzhen University, Shenzhen, China\\
\email{nidong@szu.edu.cn}\\
\and
School of Biomedical Engineering, Shenzhen University, Shenzhen, China
\and
School of Biomedical Engineering and Informatics, Nanjing Medical University, Nanjing, China
\and
Centre for Artificial Intelligence and Robotics, Hong Kong Institute of Science \& Innovation, Chinese Academy of Sciences, Hong Kong, China 
\and
Shandong Provincial Maternal and Child Health Care Hospital Affiliated to Qingdao University, Jinan, China
\and
Jiangxi Maternal and Child Health Hospital, Nanchang, China\\
\email{jxsfbcsk@163.com}\\
\and
School of Artificial Intelligence, Shenzhen University, Shenzhen, China
}
  
\maketitle              

\begin{abstract}
Accurate fetal birth weight (FBW) estimation shortly before delivery is clinically valuable yet challenging due to its reliance on operator expertise, particularly in low-resource settings.
To reduce this reliance, we study near-term birth-weight regression from blind-sweep ultrasound (US) videos acquired within 48 hours prior to delivery, with post-delivery weighing as ground truth.
Accordingly, we propose a foundation model–driven key anatomy frame selection framework that enables accurate FBW regression despite the absence of plane constraints in blind sweeps.
Our highlights are as follows:
(1) We believe this is the first work to estimate FBW using blind-sweep US videos, enabling operator-independent assessment.
(2) An Anatomy-Guided Frame Selection module equipped with a vision-language foundation model is proposed for keyframe collection in unconstrained sweeps.
(3) A Redundancy-Aware Feature Compression module is designed to compress frame features while preserving task-relevant information, alleviating temporal redundancy.
Extensively validated on prospectively collected data from 839 patients, our method achieves an MAE of 161.3 g, with 90.23\% and 100\% of cases falling within 10\% and 15\% absolute percentage error, outperforming typical Hadlock estimation and strong competitors. Codes are available at \url{https://github.com/ouleoule/BlindSweep-EBW}.

\end{abstract}

%
%
%

\section{Introduction}

Accurate estimation of fetal birth weight (FBW) prior to delivery is crucial for determining the mode of delivery and managing perinatal risks. 
Ultrasound (US) remains the primary modality for estimation of fetal weight in clinical practice~\cite{hadlock1985efw_prospective}.

However, in real-world clinical settings, accurate FBW estimation requires obtaining standard planes and performing precise biometric measurements~\cite{Salomon2019ISUOGBiometryGrowth}. This process is highly dependent on operator expertise. Such dependency is particularly problematic in low-resource setting, where the shortage of experienced sonographers and the high demand for obstetric services further exacerbate assessment inaccuracies and potential clinical risks.

Recent artificial intelligence (AI) has advanced rapidly in US analysis~\cite{zhang2026artificial}, demonstrating strong performance in plane localization, segmentation, and biometric measurement~\cite{yang2021searching,huang2025uncertainty,venturini2025wholeexamai,huang2022online,ramirezzegarra2023ai_fetal_ultrasound}.
P\l{}otka et al.~\cite{plotka2023efw_abdomen_ultrasound,plotka2025direct_biometry_video} proposed a deep learning framework to extract key features from fetal abdominal images for real-time weight estimation. Wang et al.~\cite{wang2025accurateefficientfetalbirth} utilized 3D ultrasound data to capture richer anatomical information, significantly improving accuracy under challenging fetal postures.
These studies have demonstrated strong performance in fetal weight (FW) estimation. However, they still rely heavily on high-quality, standardized input data. When operators have limited experience, suboptimal image acquisition can lead to substantial degradation in prediction accuracy. Hence, current methods have not fundamentally addressed the issue of operator dependency.

This tight coupling between performance and standardized operation limits the scalability of AI in primary care.
To address this challenge, Stringer et al.~\cite{gomes2022communmed_mobileai_ga_malpresentation,pokaprakarn2022nejmevidence_blindsweeps,viswanathan2024ijgo_flyto_cineloop_qc,stringer2024jama_blindsweeps} proposed replacing the conventional plane-based paradigm with blind-sweep US, which acquires data through continuous scanning along predefined trajectories without requiring real-time identification of standard anatomical structures.

Tanya Akumu et al.~\cite{akumu2025adaptiveframesel} proposed a clustering-based strategy to adaptively select informative frames from blind-sweep videos, effectively reducing redundancy and enhancing the robustness of gestational age estimation.
These studies have demonstrated that blind-sweep US can reduce dependence on operator expertise while maintaining competitive performance.

In this study, we propose a foundation model-driven key anatomy frame selection framework for blind-sweep US FBW estimation.
Our contributions are threefold.
(1) To the best of our knowledge, this is the first study to apply the blind-sweep paradigm to FBW estimation, effectively decoupling accuracy from the need for expert-level image acquisition.
(2) We propose an Anatomy-Guided Frame Selection (AFS) module that leverages a foundation model and anatomical priors to implicitly localize key structures within unconstrained blind-sweep sequences.
(3) We introduce a Redundancy-Aware Feature Compression (RAFC) module that compresses frame features while enhancing task-relevant information, mitigating the impact of temporal redundancy in blind-sweep videos.
We prospectively collected a dataset of 870 exams from 839 patients.
Experimental results demonstrate the effectiveness of our method, which reduces operator dependency without requiring high-quality input data, providing reliable FBW estimation in resource-constrained settings.

\begin{figure}[!t]
  \centering
  \includegraphics[width=0.95\textwidth]{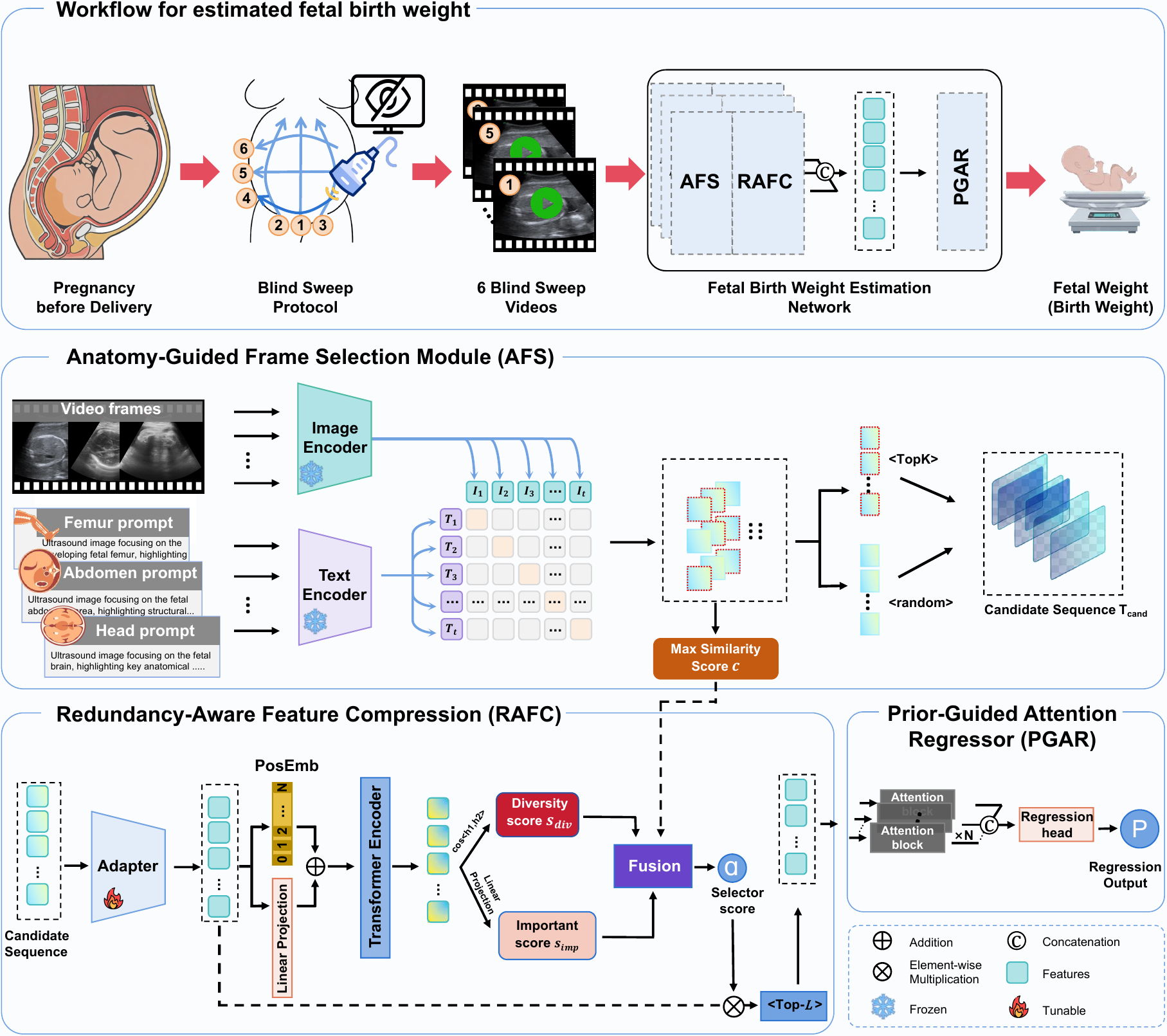}
  \caption{The overview of our proposed framework.}
  \label{fig:model-illustration}
\end{figure}

\section{Methodology}
Blind-sweep US videos exhibit variable temporal lengths, rapid viewpoint transitions, sparse standard anatomical planes, and substantial acquisition noise. Directly regressing FBW from all frames introduces significant redundancy and may dilute discriminative anatomical cues.
To address this challenge, we propose a hierarchical anatomy-guided frame selection framework for blind-sweep FBW estimation (Fig.~\ref{fig:model-illustration}). Without relying on explicit standard plane detection, our method progressively refines informative evidence through two selection stages, followed by prior-guided regression. Specifically, the framework consists of: (1) an AFS module that constructs a coarse informative frame subset using anatomical priors; (2) a RAFC module that suppresses temporally redundant frames and enhances task-relevant representations; and (3) a Prior-Guided Attention Regressor (PGAR) integrating refined frame features for FBW estimation.

\subsection{Anatomy-Guided Frame Selection}
Blind-sweep US videos inherently contain noise-dominant or low-quality frames, which may obscure discriminative anatomical structures and degrade regression performance. To preserve frames with vital anatomical content, we propose an AFS module that leverages anatomical priors to identify informative frames.

We adopt FetalCLIP~\cite{maani2026fetalclip} as the backbone encoder due to its strong cross-modal representation capability learned from large-scale fetal image-text alignment. It consists of an image encoder $\mathcal{E}_{img}$ and a text encoder $\mathcal{E}_{txt}$ that project visual frames and anatomical concept descriptions into a shared embedding space.

Given a blind-sweep video $V = \{v_1, \dots, v_N\}$ and a anatomical concepts set $T = \{t_1, \dots, t_M\}$, where each concept $t_M$ is described using the predefined and fixed text prompts provided in the public FetalCLIP repository.
we extract embeddings for the videos and concepts set as:
\begin{equation}
    X = \mathcal{E}_{img}(V),
\end{equation}
\begin{equation}
    P = \mathcal{E}_{txt}(T),
\end{equation}
where $N$ represents the number of video frames and $M$ represents the number of concepts. The anatomical relevance of each frame is quantified by the cosine similarity between frame and concept embeddings $ \mathbf{A} = cos(X, P) $.
We define the anatomical confidence of each frame as the maximum similarity over all concepts:
\begin{equation}
    c = \max_{j} \mathbf{A}_{:,j} \in \mathbb{R}^{N}.
\end{equation}

Frames with higher confidence are more likely to contain informative anatomical structures. The top-$K$ frames are selected according to $c$ to form the coarse informative frame subset:
\begin{equation}
    \mathcal{I} = \operatorname{TopK}(c), \quad X_c = X[\mathcal{I}, :].
\end{equation}
This selection effectively filters out noise-dominant frames while preserving structurally meaningful evidence for subsequent redundancy-aware refinement and regression.

\subsection{Redundancy-Aware Feature Compression}
Blind-sweep videos often contain many temporally adjacent frames with high visual similarity, which can dilute the model's attention and reduce the discriminative power of frame representations. 
To address this, we propose a Redundancy-Aware Feature Compression (RAFC) module that compresses frame features while enhancing task-relevant information.

Given $X_c$ obtained from the AFS module, we first adapt the features to the fetal weight estimation task using a lightweight adapter:
\begin{equation}
    \tilde{X} = \text{Adapter}(X_c).
\end{equation}

Next, a transformer layer is applied to model inter-frame relationships and learn frame-level importance: $ H = \text{Transformer}(\tilde{X})$, H is the K x D feature matrix.
To explicitly estimate the importance of each frame, we compute a task-specific importance score via a linear mapping followed by a sigmoid activation $\sigma$:
\begin{equation}
    s_{imp} = \sigma(W_{\text{imp}} H), s_{\mathrm{imp}}\in\mathbb{R}^{K}
\end{equation}

To account for temporal redundancy after AFS candidate selection, we compute a local diversity score for each candidate frame within a sliding window of length $w$:
\begin{equation}
    d_{i} = 1 - \max_{j \in \mathcal{W}_i} cos(h_i, h_j), \quad i=1,\dots,K,
\end{equation}
where $\mathcal{W}_i$ denotes the set of neighboring frames within the window centered at frame $i$. Frames that are highly similar to their neighbors have lower $d_i$, indicating higher redundancy. 
For the entire video, we denote local diversity scores as $s_{\text{div}} = [d_1, d_2, \dots, d_K] \in \mathbb{R}^K$.

Finally, we adaptively fuse $c$ from AFS, frame importance $s_{\text{imp}}$, and redundancy $s_{\text{div}}$ to obtain a selector score for each frame:
\begin{equation}
    s_i = \sigma \Big(W_g [s_{\text{imp},i}, s_{\text{div},i}, c_i] \Big), \quad i=1,\dots,K,
\end{equation}
The top-$L$ frames with the highest selector scores are then retained as the fine-grained frame subset $H_f$ for subsequent regression.

\subsection{Prior-Guided Attention Regressor}
Given the fine-grained frame features $H_f$ retained after redundancy-aware compression, along with their corresponding frame scores $a$, we perform prior-guided feature aggregation for final fetal weight estimation.

The frame scores are first converted into a normalized prior distribution:
\begin{equation}
    p_i = softmax(\frac{a_i}{\tau}),
\end{equation}
where $\tau$ is a temperature parameter controlling distribution sharpness.
Instead of directly applying attention on the original features, we first re-weight the frame representations using the prior distribution:
$\tilde{H}_f = p \times H_f$, 
where $\times$ denotes row-wise scaling of frame features.

The re-weighted features are then fed into an attention module to model inter-frame dependencies:
\begin{equation}
H_{att} = \text{Attention}(\tilde{H}_f).
\end{equation}

Finally, the global average pooling will be applied to obtain a compact representation $z$, which is then passed through a regression head to predict the FBW, with the formula as: $\hat{y} = \mathcal{R}(z)$.
This prior-guided re-weighting strategy can ensure that anatomically relevant and non-redundant frames contribute more prominently during relational modeling and regression.

\section{Experiments and Results}

\textbf{Datasets and Implementations.}
With the Institutional Review Board (IRB) approval, we prospectively collected 870 ultrasound exams from 839 pregnancies, including cases with up to two eligible scans per pregnancy. All scans were acquired by 10 trained sonographers and performed within 48 hours prior to delivery using a GE LOGIQ P9 system. A standardized blind-sweep protocol was adopted, consisting of a fixed number of vertical and horizontal freehand sweeps across the maternal abdomen, yielding six blind-sweep loop video per exam. Each loop video was set to 10 seconds. Prior to delivery, the estimated fetal weight (EFW) was determined by a senior doctor using the Hadlock formula based on manual biometric measurements. Post delivery, the actual birth weight (FBW) was recorded as the gold standard for this study. The mean FBW in this study was $3282.47 \pm 400.27$ g, ranging from 1410 g to 4360 g. The dataset was split into training, validation, and testing sets at the pregnancy level with a ratio of 7:1:2 to avoid data leakage. Detailed statistics of the dataset are provided in Table~\ref{tab1}.

Prior to being fed into the network, all ultrasound frames were resized to $224 \times 224$ pixels. In the hierarchical selection process, the number of coarse candidate frames $K$ in the AFS module was set to 100. Subsequently, the PAFC module selected $L=50$ fine-grained frames from this candidate pool, with the sliding-window length set to $w=10$. For each exam, the six blind-sweep loops were processed independently, and the $L$ frame features selected from each loop were concatenated into $6L$ features for a single FBW prediction. These selection hyperparameters were determined through validation-set sensitivity analyses. To enhance the model's generalizability and robustness, online data augmentation was applied during training, including random rotations, horizontal/vertical flipping, and adjustments to contrast and brightness. We trained the model for 100 epochs with a batch size of 4 using the Adam optimizer with an initial learning rate of $5e^{-4}$, which followed a cosine annealing decay schedule. To prevent overfitting, an early stopping strategy was incorporated with a patience of 5 epochs based on the validation loss. All experiments were conducted using PyTorch, and the models were trained on an NVIDIA H20 GPU with 141 GB memory.

For evaluation metrics, we used Mean Absolute Error (MAE) and Mean Absolute Percentage Error (MAPE) to evaluate the performance. In addition, we calculated the percentage of cases with absolute percentage error within the clinically relevant thresholds of 10\% and 15\% \cite{Salomon2019ISUOGBiometryGrowth} (denoted as PC10 and PC15).

\begin{table}[!t]
\centering  
\caption{Dataset statistics and split protocol.}
\label{tab1}

\begin{tabular}{lcccccc}
\hline
Patients  &  Exams  &  Videos  &  Frames/video  &  Resolution  &  Train/Val/Test (exams)  \\
\hline
839  &  870  &  5220  &  137--588  &  $224\times224$  &  609 / 87 / 174  \\
\hline
\end{tabular}
\end{table}

\begin{table}[!t]
\centering
\caption{Test-set performance for near-term blind-sweep birth weight estimation. }
\label{tab2}
\small 
\begin{tabular}{lccccc}
\hline
Method & MAE (g)  & MAPE (\%)  & $\leq 10\%$ (\%)  & $\leq 15\%$ (\%)  & Std. (\%)  \\
\hline
ResNet50+WAA\cite{pokaprakarn2022nejmevidence_blindsweeps} & 309.50 & 9.15 & 58.15 & 79.11 & 7.09 \\
SelectGA~\cite{akumu2025adaptiveframesel}   & 287.03 & 8.49 & 65.69 & 84.52 & 6.87 \\
USFM~\cite{USFM}         & 247.26 & 6.67 & 79.60 & 91.54 & 6.25 \\
EchoNet~\cite{EchoNet}      & 295.72 & 8.90 & 69.75 & 82.01 & 6.79 \\
ViFi-CLIP~\cite{ViFi-CLIP}    & 284.94 & 8.62 & 63.79 & 84.48 & 6.57 \\
\hline
INTERGROWTH-21st~\cite{Stirnemann2017INTERGROWTH21EFW} & 216.63 & 6.48 & 77.59 & 94.83 & 4.34 \\
Hadlock-4~\cite{hadlock1985efw_prospective}    & 183.57 & 5.60 & 82.18 & 97.13 & 4.13 \\
\textbf{Ours} & \textbf{161.34} & \textbf{4.89} & \textbf{90.23} & \textbf{100.0} & \textbf{3.47} \\
\hline
\end{tabular}
\end{table}

\begin{table}[!t]
\centering
\caption{Ablation results of the proposed components. $\downarrow$ ($\uparrow$) indicates lower (higher) values are preferred.}
\label{tab:ablation}
\small 
\begin{tabular}{ccccccc}
\hline
ASF & RAFC & MAE (g) $\downarrow$ & Std. (\%) $\downarrow$ & $\leq 10\%$ (\%) $\uparrow$ & $\leq 15\%$ (\%) $\uparrow$ & MAPE (\%) $\downarrow$ \\
\hline
$\times$ & $\times$ & 253.04 & 5.91 & 72.41 & 89.08 & 6.82 \\
$\checkmark$ & $\times$ & 205.49 & 4.66 & 80.02 & 94.17 & 6.29 \\
$\times$ & $\checkmark$ & 183.54 & 4.96 & 83.62 & 93.79 & 5.73 \\
$\checkmark$ & $\checkmark$ & \textbf{161.34} & \textbf{3.47} & \textbf{90.23} & \textbf{100.00} & \textbf{4.89} \\
\hline
\end{tabular}
\end{table}

\begin{figure}[!t]
  \centering
  \includegraphics[width=\textwidth]{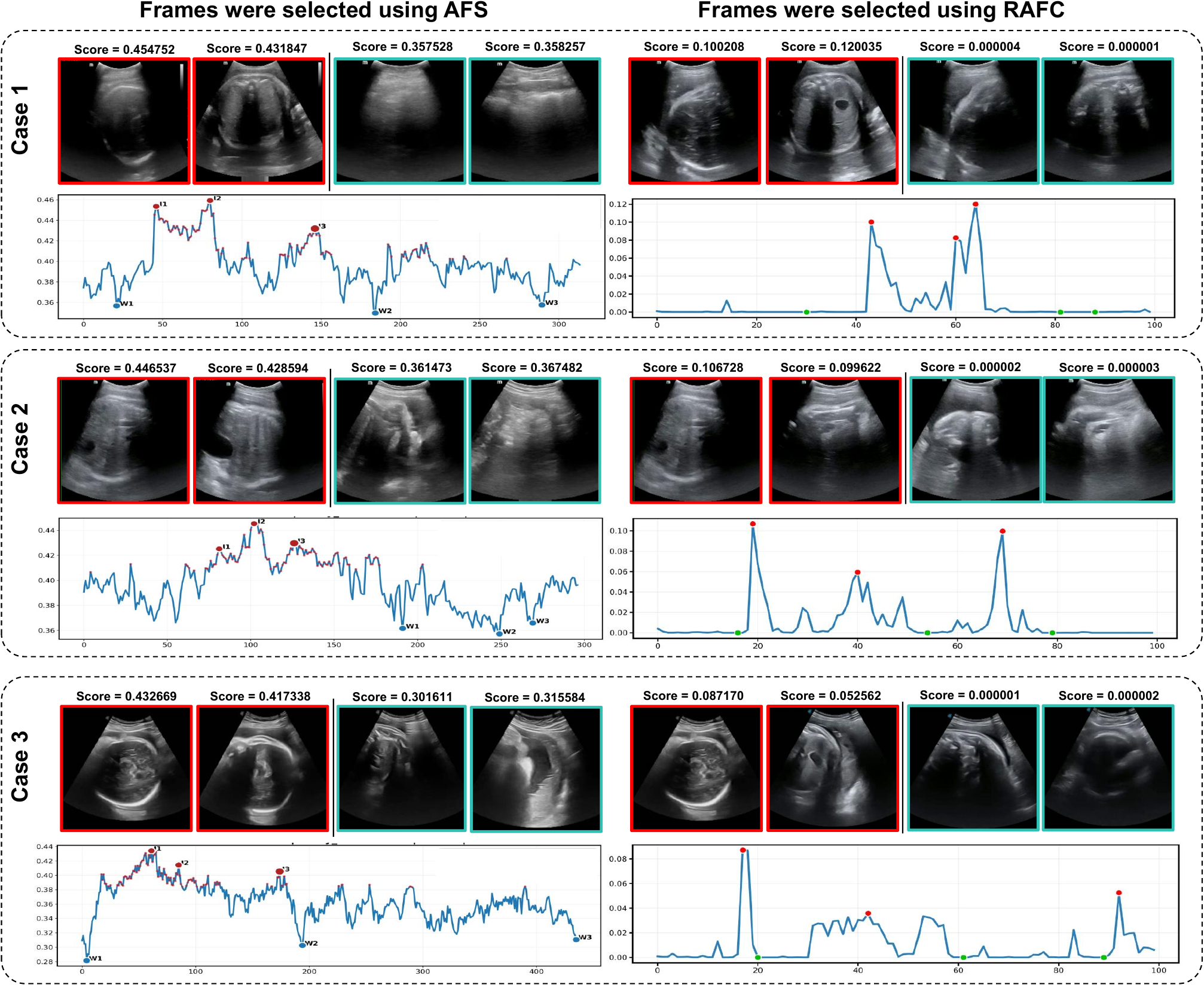}
  \caption{\textbf{Visualization of frame selection for 3 representative cases using the AFS and RAFC modules}. For each case, the line plot shows the selection score over the frame sequence (x-axis: frame index, y-axis: selection score). The top images display the video frames corresponding to the highest (red boxes) and lowest (green boxes) scores selected by each method.}
  \label{fig:show}
\end{figure}

\textbf{Quantitative Analysis.}
We comperhensively compared our approach with several competitor methods, including the clinical Hadlock-4~\cite{hadlock1985efw_prospective} (based on senior sonographers’ manual biometry), INTERGROWTH-21st~\cite{Stirnemann2017INTERGROWTH21EFW}, the blind-sweep-specific SelectGA~\cite{akumu2025adaptiveframesel} (re-purposed for FBW estimation), and several deep learning architectures, including USFM~\cite{USFM}, EchoNet~\cite{EchoNet}, and ViFi-CLIP~\cite{ViFi-CLIP}. For a fair comparison, all deep learning baselines used their original backbones with their final layers adapted to our regression head for task-specific optimization.

As shown in Table~\ref{tab2}, our method achieved the best performance with an MAE of 161.34 g, a MAPE of 4.89\%, and PC10/PC15 values of 90.23\% and 100\%, respectively. Notably, it slightly outperformed the clinical Hadlock-4 benchmark, demonstrating its competitive advantage.
Although USFM is pre-trained on a large number of standard ultrasound planes and shows advantages in related tasks, it fails to reach the Hadlock baseline in our blind-sweep scenario, where standard planes are scarce. Given that a PC10 of 80\% and PC15 of 95\% are generally considered thresholds for clinical acceptance, our method is the only one to meet these practical requirements, highlighting its potential for real-world deployment.
Ablation experiments were conducted to quantify the contribution of each module, as shown in Table~\ref{tab:ablation}. ASF contributed to a 18.8\% improvement in MAE and an 10.5\% increase in PC10. RAFC contributed to a 27\% improvement in MAE and an 11.2\% increase in PC10. Combining RAFC with ASF further boosted performance by an additional 8.7\% in MAE and 6.6\% in PC10. These results demonstrate the effectiveness of our proposed modules.

\textbf{Qualitative Analysis.}
We visualize several representative cases in Fig.~\ref{fig:show}. The results reveal that our framework effectively selects structure-rich frames. Frames selected by the first-level AFS module contain key fetal anatomical structures, including the head and abdomen, while non-informative or noisy frames are filtered out. After the second-level RAFC module performs frame compression, the remaining frames preserve complete anatomical information, and low-quality images are discarded. These results demonstrate the effectiveness of our proposed hierarchical anatomy-guided frame selection strategy.

\section{Conclusion}
In this study, we propose a foundation model-driven key anatomical frame selection framework to estimate near-term FBW from blind-sweep US videos.
Our method employs a hierarchical anatomy-guided frame selection strategy to obtain more accurate FBW predictions. (1) We introduce an AFS module that leverages anatomical priors to implicitly select informative frames and construct a coarse-level subset, filtering out non-informative frames. (2) We propose a RAFC module that compresses temporally redundant frames to form a fine-grained subset, enhancing task-relevant representations. (3) We propose a PGAR to integrate frame features for accurate FBW estimation. Extensive experiments show that our approach effectively reduces operator dependency and provides accurate FBW estimation in resource-constrained settings, with the potential to markedly improve prenatal care accessibility and quality in low-resource areas.
In future work, we will perform comprehensive multi-center clinical validation and prospective evaluation.

\begin{credits}
\subsubsection{\ackname} This work was supported by the Frontier Technology Development Program of Jiangsu Province (No. BF2024078), National Natural Science Foundation of China (No. 12326619, 82260307), Science and Technology Planning Project of Guangdong Province (No. 2023A0505020002), Jiangxi Provincial Health Technology Project (No. 202610694, 202510072), Shandong Medical and Health Science and Technology Project (No. 202309020945, 202309020926), China Maternal and Child Health Association Research Project (No. 2023CAMCHS003A12, 2023CAMCHS003A05).
\subsubsection{\discintname}
The authors have no competing interests to declare that are relevant to the content of this article.
\end{credits}

\bibliographystyle{splncs04}
\bibliography{Paper-1307}
\end{document}